\def\year{2021}\relax
\documentclass[letterpaper]{article} 
\usepackage{aaai20}  
\usepackage{times}  
\usepackage{helvet} 
\usepackage{courier}  
\usepackage[hyphens]{url}  
\usepackage{graphicx} 
\usepackage{graphicx}  
\usepackage{amssymb}
\usepackage{algorithm}
\usepackage{algpseudocode}
\usepackage{multirow}
\usepackage{array}
\usepackage{subfigure}
\usepackage{amsmath}
\usepackage{booktabs}
\usepackage{amsthm}

\makeatletter
\def\hlinew#1{%
  \noalign{\ifnum0=`}\fi\hrule \@height #1 \futurelet
   \reserved@a\@xhline}
\usepackage{color}
\usepackage{bm}

\newcommand{\ldz}[1]{\textcolor{black}{#1}}

\title{F2Net: Learning to Focus on the Foreground for Unsupervised Video Object Segmentation}

\author{Daizong Liu\textsuperscript{\rm 1$^\dagger$}, Dongdong Yu\textsuperscript{\rm 2$^\dagger$}, Changhu Wang\textsuperscript{\rm 2}, Pan Zhou\textsuperscript{\rm 1*}\\
\textsuperscript{\rm 1}Huazhong University of Science and Technology\\
\textsuperscript{\rm 2}ByteDance AI Lab\\
\{dzliu, panzhou\}@hust.edu.cn, \{yudongdong, wangchanghu\}@bytedance.com
}
\begin{document}
\maketitle
\begin{abstract}
Although deep learning based methods have achieved great progress in unsupervised video object segmentation, difficult scenarios (e.g., visual similarity, occlusions, and appearance changing) are still not well-handled. To alleviate these issues, we propose a novel \textit{Focus on Foreground Network} (\textbf{F2Net}), which delves into the intra-inter frame details for the foreground objects and thus effectively improve the segmentation performance. Specifically, our proposed network consists of three main parts: Siamese Encoder Module, Center Guiding Appearance Diffusion Module, and Dynamic Information Fusion Module. Firstly, we take a siamese encoder to extract the feature representations of paired frames (reference frame and current frame). Then, a Center Guiding Appearance Diffusion Module is designed to capture the inter-frame feature (dense correspondences between reference frame and current frame), intra-frame feature (dense correspondences in current frame), and original semantic feature of current frame. Specifically, we establish a Center Prediction Branch to predict the center location of the foreground object in current frame and leverage the center point information as spatial guidance prior to enhance the inter-frame and intra-frame feature extraction, and thus the feature representation considerably focus on the foreground objects. Finally, we propose a Dynamic Information Fusion Module to automatically select relatively important features through three aforementioned different level features. Extensive experiments on DAVIS2016, Youtube-object, and FBMS datasets show that our proposed \textbf{F2Net} achieves the state-of-the-art performance with significant improvement.
\end{abstract}

\section{Introduction}
Unsupervised video object segmentation (UVOS)
aims to separate foreground objects from their background in a video sequence without any prior information. Due to the lack of prior knowledge about the foreground objects, this task not only suffers from the common challenges (e.g. object deformation and occlusion) in other video-related tasks, but also faces huge difficulties in accurately discovering the most prominent and distinct objects across video frames from a complex and diverse background. 

\begin{figure}[t!]
\centering
\includegraphics[width=0.48\textwidth]{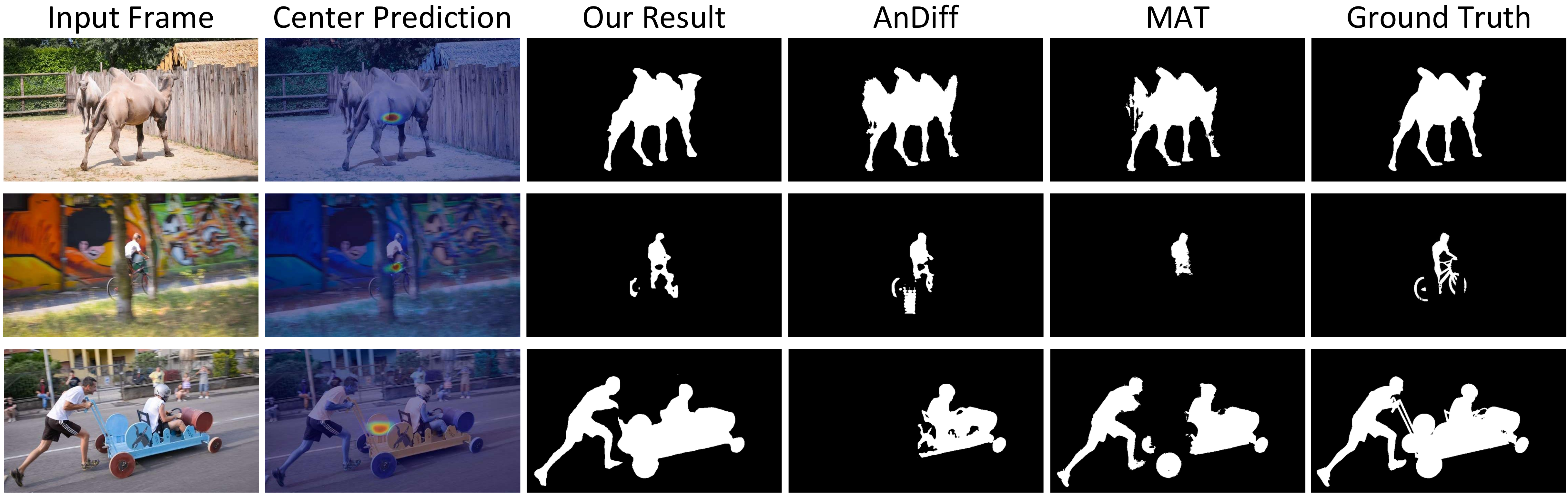}
\caption{Compared to other methods, our network shows better segmentation performance under the challenging scenarios of visual similarity, occlusion, and appearance changing by considering the center information of the foreground.}
\label{fig:introduction}
\vspace{-12pt}
\end{figure}

Traditional methods tend to address this task by using handcrafted or learnable features, such as objectness \cite{zhang2013video}, motion boundary \cite{papazoglou2013fast}, saliency \cite{wang2015saliency}, and trajectories \cite{ochs2011object}. These are typically non-learning methods working in a purely unsupervised manner without using any training data.
Recently, more research efforts have been devoted to tackling this task in deep learning frameworks,
leading to an unsupervised solution (no annotation used for any testing frame).
Many of these \cite{zhou2020motion,cheng2017segflow,li2018unsupervised} employ two-stream networks to combine local motion and appearance information. Due to the usage of optical flow, they may fail to correctly infer the foreground when the object is occluded or nearly static~\cite{tokmakov2017learning}. 
Latest appearance matching based methods \cite{wang2019zero,yang2019anchor,lu2019see} are proposed to explore higher-order relationships among video frames, which obtain more optimal results from a global view by attention mechanisms. However, they lack robustness when there are visually similar object existed in the background.

For example, as shown in Figure \ref{fig:introduction}, there are three challenging scenarios: foreground-background visual similarity, object occlusion, and appearance changing. For appearance matching method Anchor Diffusion Network (AnDiff) \cite{yang2019anchor}, it fails to separate the foreground camel from the background one in the first video as the two camels have similar appearance.
It also can not handle the appearance changing problem as shown in the third video. For two stream network MAT \cite{zhou2020motion}, it only segments part of the occluded object in the second video because optical flow is not robustness to the object occlusion.
The above methods pay less attention to the foreground object, leading to the inaccurate segmentation result.
Considering center point can be taken as the spatial prior guidance \cite{zhou2019objects,zhou2020tracking,wang2020centermask}, we tend to firstly predict the center point of the primary object and then segment the mask from such point to its surroundings. Therefore, our network can focus more on the foreground object, and alleviate the visually similarity, occlusion and appearance changing problems.

Towards this end, we propose a novel \textit{Focus on Foreground Network}  (\textbf{F2Net}) for unsupervised video object segmentation, which exploits center point information to focus on the foreground object. 
Different from the common appearance matching based methods, we additionally establish a Center Prediction Branch to estimate the center location of the primary object.
Then, we encode the predicted center point into a gauss map as the spatial guidance prior to enhance the intra-frame and inter-frame feature matching in our Center Guiding Appearance Diffusion Module, leading the model to focus on the foreground object. 
After the appearance matching process, we can get three kinds of information flows: inter-frame features, intra-frame features, and original semantic features of current frame.
Instead of fusing these three features by simple concatenation like previous methods, an attention based Dynamic Information Fusion Module is developed to automatically select the most discriminative features across the three features, providing more optimal representations for final segmentation.

To summarize, our main contributions are three-folds:
\begin{itemize}
    \item To the best of our knowledge, we are the first to take center point information into UVOS task for spatial guidance prior, which helps model focus on the foreground object. Specifically, our proposed Center Guiding Appearance Diffusion Module leverages the center prior in appearance matching procedure to extract foreground attentive representations.
    \item We develop a Dynamic Information Fusion Module to aggregate information from different level features, which generates more discriminative representations for the final foreground object segmentation.
    \item Extensive experiments are conducted on three popular UVOS benchmarks, DAVIS2016, FBMS, and Youtube-Objects. Compared to the state-of-the-art methods, we achieve the significant improvement with a large margin.
\end{itemize}

\section{Related Work}
\noindent \textbf{Unsupervised Video Object Segmentation.}
UVOS aims to automatically separate foreground object(s) from
their background in a video without any human intervention. 
Early methods typically utilize handcrafted features (e.g., color, optical flow) \cite{papazoglou2013fast,faktor2014video,tsai2016video,hu2018unsupervised}.
Recently, benefiting from the large datasets \cite{perazzi2016benchmark}, more research efforts have been devoted to tackling this task in deep learning frameworks.
Tokmakov et al. \cite{tokmakov2017learning} proposed
a purely optical flow based network that discards appearance modelling and casts segmentation as foreground motion prediction,
thus poorly deals with static foreground objects. To address this problem, two-stream networks are introduced to fuse appearance and motion information \cite{li2018flow,jain2017fusionseg,cheng2017segflow,li2018unsupervised,zhou2020motion}. However, above methods utilize optical flow information, and significantly suffer from not only the large computation of the optical flows, but also the deterioration in the quality of their predictions over time. 
Targeting this issue, several approaches \ldz{\cite{wang2019zero,yang2019anchor,lu2019see,chen2018blazingly,fathi2017semantic,li2018instance,oh2019video}} tackle video object segmentation by simply learning similarities between pixel embeddings without motion contexts. AGNN \cite{wang2019zero} provides an unified, end-to-end trainable network to capture the higher-order correlated information with graph attention network. AnDiff \cite{yang2019anchor} performs appearance similarity learning, feature propagation and binary segmentation in a single network. COS \cite{lu2019see} utilizes co-attention to comprehensively use the rich, inherent correlation information within videos.

Although these appearance matching based methods achieve state-of-the-arts performance, 
they only consider the matched pixels across the frame where the visual similar objects or surroundings may be wrongly taken as the foreground. To tackle this issue, we propose a Center Guiding Appearance Diffusion Module to match the pixel embeddings with the guidance of the center point information. It helps our model focus more on the pixels of actual foreground, thus filters out the background noise. In addition, it can alleviate the occlusion and appearance changing.

\begin{figure*}[t!]
\centering
\includegraphics[width=1.0\textwidth]{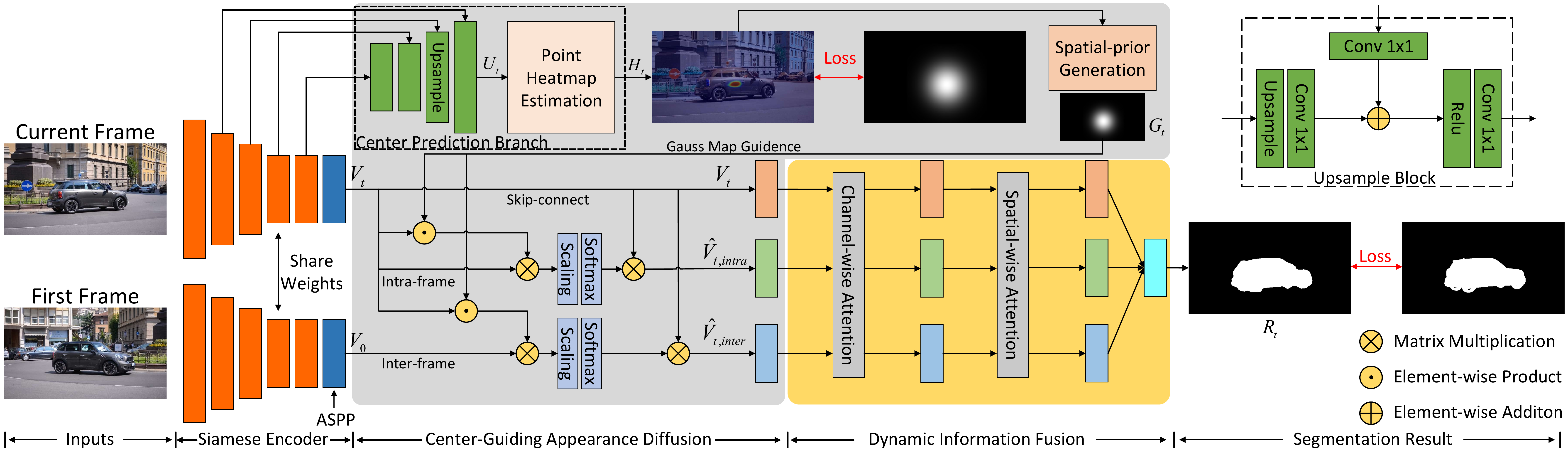}
\caption{Overall pipeline of the proposed network architecture. A pair of frames is first fed into a siamese encoder to obtain the feature representations. After that, we develop a Center Guiding Appearance Diffusion Module to first predict the center point of the foreground object in current frame, and then generate a gauss map as spatial guidance prior for the following up appearance matching procedure. At last, we devise a Dynamic Information Fusion Module to aggregate different level features for foreground object segmentation.}
\label{fig:pipeline}
\vspace{-10pt}
\end{figure*}

\noindent \textbf{Attention Mechanism.}
Differentiable attentions have been widely used in recent neural networks for various tasks, such as visual question answering \cite{lu2016hierarchical}, human pose estimation \cite{chu2017multi,su2019multi}, and image classification \cite{hu2018squeeze,li2019selective}. It allows networks to focus on the most informative parts of the inputs. Latest appearance matching methods in UVOS \cite{wang2019zero,yang2019anchor,lu2019see} generally match the pixel embeddings across the whole frame based on attention mechanism, thus may wrongly segment the visually similar backgrounds. They also fuse the matched features by simple concatenation which may lose fine-grained details. To pay more attention on the foreground, we predict the foreground object center and encode it to a spatial gauss map. Such spatial guidance is injected into the attention mechanism to selectively focus on the foreground pixels. Besides, instead of fusing features by simply concatenation, we devise an attention based Dynamic Information Fusion Module to aggregate information from different level features.

\section{Method}
\subsection{Overview}
UVOS aims to automatically segment the primary object(s) from videos. To focus on the foreground and neglect the background, we propose the \textit{Focus on Foreground Network} \textbf{F2Net} illustrated in Figure \ref{fig:pipeline}, which consists of three main parts: a Siamese Encoder Module, a Center Guiding Appearance Diffusion Module, and a Dynamic Information Fusion Module.
Given a pair of frames, we first generate their embeddings by a siamese encoder. After that, the Center Guiding Appearance Diffusion Module first predicts the foreground object center point by a Center Prediction Branch, then encodes it into a gauss map as the spatial-prior condition during the appearance matching process. After appearance matching, we can get three kinds of features. At last, we apply a Dynamic Information Fusion Module to fuse different level matched features for final segmentation.


\begin{figure}[t!]
\centering
\includegraphics[width=0.45\textwidth]{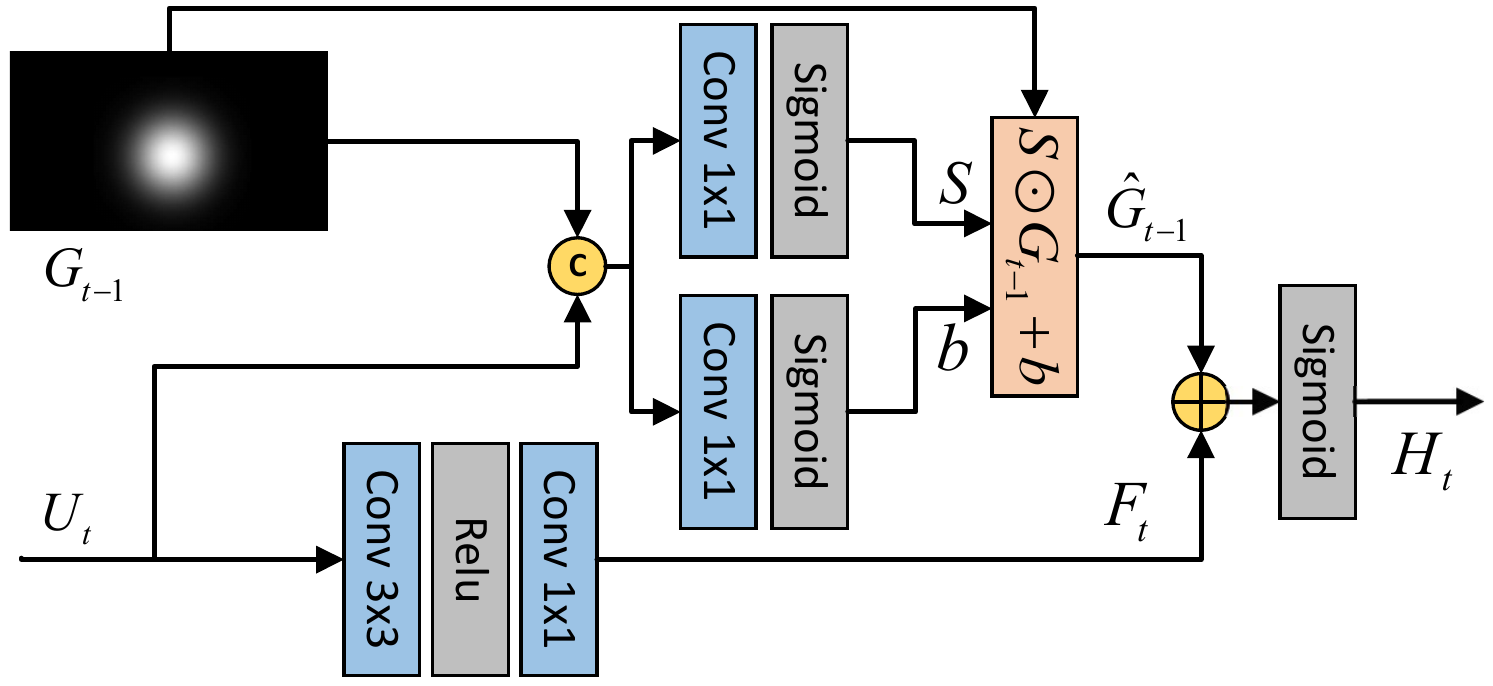}
\caption{Illustration of the heatmap estimation in current frame, which considers 1) \ldz{spatial} clues $\bm{G}_{t-1}$ from previous frame, and 2) semantic information $\bm{U}_t$ from current frame.}
\label{fig:center}
\vspace{-12pt}
\end{figure}

\subsection{Siamese Encoder}
The siamese encoder takes a pair of RGB images as inputs, including a current frame $\bm{I}_t \in \mathbb{R}^{W\times H \times 3}$ and a reference frame $\bm{I}_0 \in \mathbb{R}^{W\times H \times 3}$.
Here, we take the first frame $\bm{I}_0$ as the reference frame because it is guaranteed to contain the foreground objects.
The backbone network of the siamese encoder is DeepLabv3 \cite{chen2017rethinking}, which consists of five convolution blocks (\textit{res1}, \textit{res2}, \textit{res3}, \textit{res4}, \textit{res5}) from ResNet \cite{he2016deep} and an atrous spatial pyramid pooling (ASPP) module. We denote the extracted embeddings of $\bm{I}_t,\bm{I}_0$ as $\bm{V}_t,\bm{V}_0 \in \mathbb{R}^{\frac{W}{8}\times \frac{H}{8} \times C}$, where $W,H,C$ are the width, height, and the number of channels, respectively.

\subsection{Center Guiding Appearance Diffusion}
To focus more on the foreground object, we delve into the intra-inter features with center point information. In detail, we build a Center Prediction Branch to detect the object center point, and then utilize it as spatial guidance prior for further Spatial-Prior Guided Appearance Matching.

\noindent \textbf{Center Prediction Branch.}
This branch tends to predict the center point of the foreground object. Specially, following \cite{tompson2014joint}, we transform the point prediction into a heatmap $\bm{H}_t$ estimation task, which consists of two main steps: \textit{Feature Upsampling} and \textit{Heatmap Generation}.
In \textit{Feature Upsampling}, we adopt an upsample module to efficiently merge features of different res-blocks (\textit{res2}, \textit{res3}, \textit{res4}, \textit{res5}) in different scales to enhance the information for high-resolution features. 
As shown in Figure \ref{fig:pipeline}, we illustrate the detailed structure of upsample block. The input of previous layer is first upsampled to the same size to the skip-connected features, and then both inputs are element-wise added. The final output of the whole upsampling blocks can be denoted as $\bm{U}_t \in \mathbb{R}^{\frac{W}{4} \times \frac{H}{4} \times D}$, where $D$ is the channel number. In \textit{Heatmap Generation}, to predict the final point-wise heatmap $\bm{H}_t \in \mathbb{R}^{\frac{W}{4} \times \frac{H}{4} \times 1}$, we consider two aspects as shown in Figure \ref{fig:center}: 1) \ldz{Spatial} clues from previous frame: we propagate previous gauss map $\bm{G}_{t-1}$ to current frame for better locating the foreground
(if the current frame is the first frame, we utilize a zero map as the previous gauss map). We concatenate the current embedding $\bm{U}_t$ and the \ldz{spatial} clues $\bm{G}_{t-1}$ to learn $\bm{S}, \bm{b} \in \mathbb{R}^{\frac{W}{4} \times \frac{H}{4} \times 1}$, which are scale and bias parameters \ldz{\cite{yang2018efficient}} to control the weight to adjust the $\bm{G}_{t-1}$ under the guidance of the appearance information of current frame.
We can formulate such process as:
\begin{equation}
    \bm{S} = \text{Conv2d}(\text{Concat}[\bm{U}_t,\bm{G}_{t-1}]),
\end{equation}
\begin{equation}
    \bm{b} = \text{Conv2d}(\text{Concat}[\bm{U}_t,\bm{G}_{t-1}]),
\end{equation}
\begin{equation}
    \hat{\bm{G}}_{t-1} = \text{Sigmoid}(\bm{S}) \odot \bm{G}_{t-1} + \text{Sigmoid}(\bm{b}),
\end{equation}
where $\odot$ denotes the element-wise product. 2) Semantic information from current feature: Since $\bm{U}_t$ also contains enough information to predict the target center point, we can directly estimate the heatmap $\bm{F}_t$ on $\bm{U}_t$ by applying a $3\times 3$ convolutional layer with ReLU, followed by a $1\times 1$ convolutional layer. 
At last, we add this two estimation branch into one with a sigmoid function by:
\begin{equation}
    \bm{H}_t = \text{Sigmoid}(\hat{\bm{G}}_{t-1} + \bm{F}_t).
    \label{eq:heatmap}
\end{equation}

To choose the best center point $o_t=(x_t,y_t) \in \mathbb{R}^2$ from $\bm{H}_t$, instead of directly choosing the point with the maximum score, we additionally consider the motion clues across the sequential frames for better accuracy. We first rank top $K$ points using NMS \cite{lin2017focal} from $\bm{H}_t$ as candidates, and then utilize a motion mechanism \cite{xu2019mhp} to predict a coarse center point $p_t = p_{t-1}+\frac{1}{n}\sum_{m=t-n}^{t-1}(p_m-p_{m-1})$ using $n$ history object centers in previous frames. At last, we compute the distance from each candidate to $p_t$, and choose the closest one as the final center $o_t$ of the foreground object in current frame. In our experiments, we find it can achieve better performance than the maximum strategy. 

\noindent \textbf{Spatial-Prior Guided Appearance Matching.}
To determine the foreground object, there are two essential properties: 1) distinguishable in an individual frame
(locally saliency), and 2) frequently appearing throughout the video sequence (globally consistent).
To achieve the first goal, 
we apply a non-local operation \cite{wang2018non} on the current feature $\bm{V}_t$ to locate the salient object in an intra-wise matching way. To achieve our second goal, we utilize another non-local operation on both current and the reference features $\bm{V}_t,\bm{V}_0$ to capture the inter-frame correlated information for alleviating appearance drift. We also employ a skip-connection on current feature $\bm{V}_t$ to preserve the semantic information. All three features contain different level information as shown in Figure \ref{fig:pipeline}. To focus on the foreground, we encode the predicted center point $o_t$ into a gauss map $\bm{G}_t$ as spatial guidance prior for the intra-frame and inter-frame appearance matching. 
Compared to other appearance matching methods, the key difference of our method is that the encoded feature representation is weighted by the gauss-guided spatial prior.

As shown in Figure \ref{fig:pipeline}, 
given the intra-frame and iter-frame feature pairs $(\bm{V}_t,\bm{V}_t)$ and $(\bm{V}_0,\bm{V}_t)$, we first flatten $\bm{V}_0,\bm{V}_t$ into shape $\frac{WH}{64} \times C$, then compute the gauss-guided correlation matrices $\bm{M}_{intra},\bm{M}_{inter} \in \mathbb{R}^{\frac{WH}{64} \times \frac{WH}{64}}$ as:
\begin{equation}
    \bm{M}_{intra} = \text{Softmax}(\frac{1}{\sqrt{C}}\bm{V}_t(\bm{V}_t\odot \bm{G}_t)^T),
\end{equation}
\begin{equation}
    \bm{M}_{inter} = \text{Softmax}(\frac{1}{\sqrt{C}}\bm{V}_0(\bm{V}_t\odot \bm{G}_t)^T),
\end{equation}
where $C$ is the channel number of $\bm{V}_t$. After that, we reconstruct feature $\hat{\bm{V}}_t$ in which the pixel embeddings are weighted according to their similarity with the
foreground:
\begin{equation}
    \hat{\bm{V}}_{t,intra} = \bm{M}_{intra} \bm{V}_t,
\end{equation}
\begin{equation}
    \hat{\bm{V}}_{t,inter} = \bm{M}_{inter} \bm{V}_t.
\end{equation}
At last, we can get three kinds of information flows and reshape them back to a 3D tensor with the size of $\frac{W}{8} \times \frac{H}{8} \times C$ for further segmentation: 1) initial current feature $\bm{V}_t$ by a skip connection \cite{he2016deep}, 2) intra-frame feature $\hat{\bm{V}}_{t,intra}$ for intra-frame discriminability and 3) inter-frame feature $\hat{\bm{V}}_{t,inter}$ for inter-frame consistency.

\subsection{Dynamic Information Fusion}
Original current feature $\bm{V}_t$ only contains a coarse clue for inferring the target foreground object without a global view. Intra-frame feature $\hat{\bm{V}}_{t,intra}$ contains more accurate salient object information in current frame, but fails to address the appearance changes in a video sequence. Inter-frame feature $\hat{\bm{V}}_{t,inter}$ contains more contexts to adapt to the appearance changes of the foreground objects. Instead of directly fusing these three features by simple concatenation \cite{lu2019see,yang2019anchor}, we need to selectively aggregate them to generate more discriminative features. According to the above analysis, 
we develop a Dynamic Information Fusion Module to emphasize meaningful features along both channel and spatial dimensions with attention mechanism.

\begin{figure}[t!]
\centering
\includegraphics[width=0.5\textwidth]{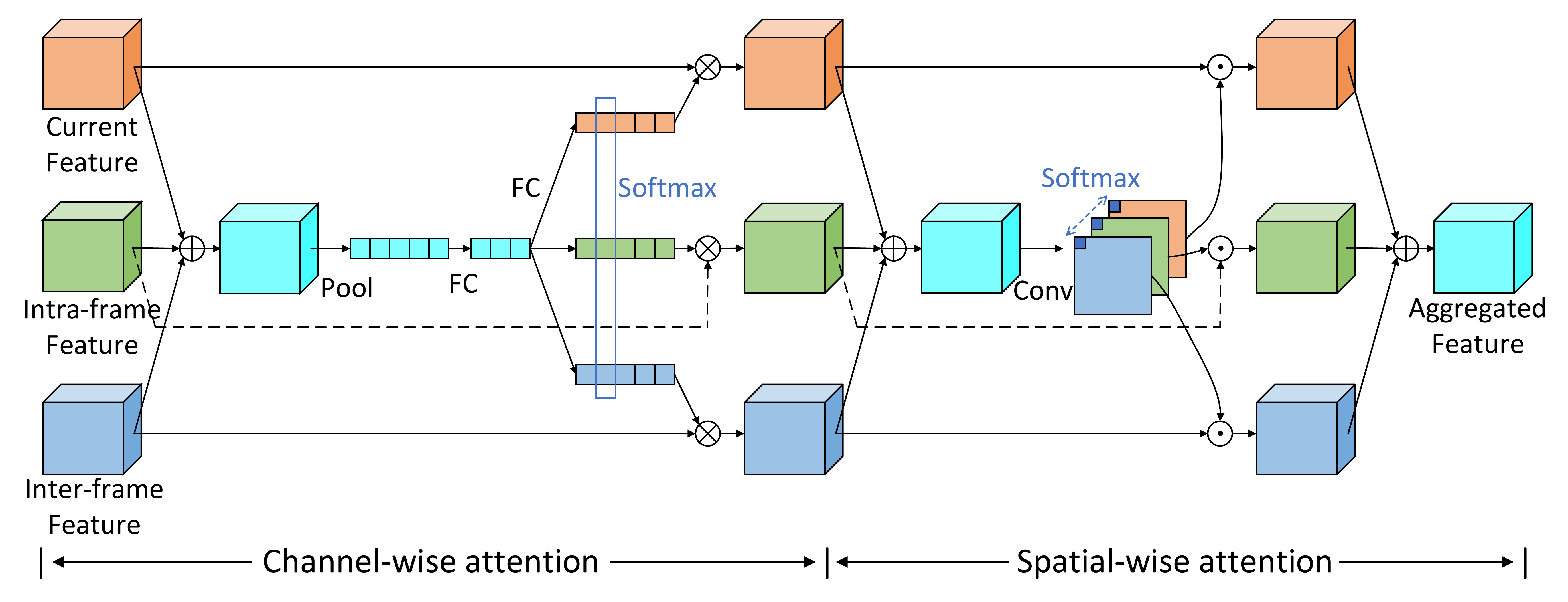}
\caption{Scheme of the Dynamic Information Fusion Module, in which we apply attentions on channel and spatial dimensions for different level features aggregation.}
\label{fig:attention}
\vspace{-12pt}
\end{figure}

\noindent \textbf{Channel-wise attention.}
The basic idea of channel-wise attention is to use gates to control the information flows from three-level features in channel dimension. To achieve this goal, the gates need to integrate information from all features. As shown in Figure \ref{fig:attention}, we first fuse three kinds of features by an element-wise addition as:
\begin{equation}
    \tilde{\bm{V}_t} = \bm{V}_t +\hat{\bm{V}}_{t,intra} +\hat{\bm{V}}_{t,inter}.
\end{equation}
Then we utilize global average pooling on $\tilde{\bm{V}_t}$ to conduct spatial squeeze, where each element is calculated by shrinking $\tilde{\bm{V}_t}$ through spatial dimensions. We further apply a simple fully connected (FC) on the squeezed feature to enable the guidance for adaptive selection. After that, we employ another three FC layers to generate the gate vectors $\{\bm{Z}_1,\bm{Z}_2,\bm{Z}_3\} \in \mathbb{R}^{1 \times 1 \times C}$ for $\{\bm{V}_t,\hat{\bm{V}}_{t,intra},\hat{\bm{V}}_{t,inter}\}$, respectively, and utilize
channel-wise softmax function to generate adaptive weights $\{\bm{W}_1,\bm{W}_2,\bm{W}_3\}  \in \mathbb{R}^{1 \times 1 \times C}$ corresponding to different level features $\{\bm{Z}_1,\bm{Z}_2,\bm{Z}_3\}$ as:
\begin{equation}
    \bm{W}^c_i = \frac{exp(\bm{Z}_i^c)}{\sum_{j=1}^3 exp(\bm{Z}_j^c)}, \quad i \in \{1,2,3\},
\end{equation}
in which, $\bm{W}^c_i$ represents relative importance of feature $\bm{Z}_i$ at channel $c \in C$, and $\sum_{i=1}^3\bm{W}^c_i=1$.
The gated feature maps for three information flows can be formulated as:
\begin{equation}
\begin{aligned}
    \bm{V}_t = \bm{V}_t \odot & \bm{W}_1, \quad
    \hat{\bm{V}}_{t,intra} = \hat{\bm{V}}_{t,intra} \odot \bm{W}_2, \\
    &\hat{\bm{V}}_{t,inter} = \hat{\bm{V}}_{t,inter} \odot \bm{W}_3.
\end{aligned}
\end{equation}

\noindent \textbf{Spatial-wise attention.} Similar to the channel-wise attention, we also first fuse three features by element-wise addition. To strengthen the information on the spatial dimension, we then employ $1 \times 1$ convolutional filters to conduct the channel squeeze and compress the channels to 3, where the feature map of each channel corresponding to the spatial weights for each level feature. After that, a softmax operation is conducted across channels to rescale activations and the rescaled activations are sliced along the channel dimension to generate pixel-wise adaptive weights. At last, we sum the three features to one according to the generated weights.

After the channel-wise and spatial-wise attention modules, the information flows from three-level input features can be adaptively aggregated into one feature map. Like most of previous works, we employ a decoder which consists of two convolutional layers to generate the final binary mask result $\bm{R}_t \in \mathbb{R}^{W\times H}$.

\subsection{Implementation Details}
\noindent \textbf{Training loss.}
Given a pair of input images, our \textbf{F2Net} predicts a center point heatmap $\bm{H}$ and a mask result $\bm{R}$ of the current frame.
For point heatmap estimation, we apply an element-wise focal loss \cite{lin2017focal} on the heatmap $\bm{H}$ in Eq.(4) and the corresponding ground-truth gauss map $\hat{\bm{H}}$:
\begin{equation}
    \mathcal{L}_{f} = \sum_{(x,y)}\left\{  
             \begin{array}{l}
             (1-H_{(x,y)})^\alpha \text{log}(H_{(x,y)}),\  \text{if}\ \hat{H}_{(x,y)}=1   \\
             (1-\hat{H}_{(x,y)})^\beta (H_{(x,y)})^\alpha \text{log}(1-H_{(x,y)}),\  \text{else} \\  
             \end{array}
\right. 
\end{equation}
where $H_{(x,y)}$ is the score at location $(x, y)$ in the predicted heatmap $\bm{H}$, and we set $\alpha$ as 2 and $\beta$ as 4 following the default setting in \cite{law2018cornernet}. For segmentation mask result, we employ the binary cross entropy loss on the predict mask $\bm{R}$ and the ground truth $\hat{\bm{R}}$ as:
\begin{equation}
    \mathcal{L}_{b} = -\sum_{(x,y)}\hat{R}_{(x,y)}\text{log}(R_{(x,y)})+(1-\hat{R}_{(x,y)})\text{log}(1-R_{(x,y)}).
\end{equation}
The overall loss function can be obtained by:
\begin{equation}
    \mathcal{L} = \mathcal{L}_{f}+\mathcal{L}_{b}.
\end{equation}

\noindent \textbf{Training settings.} Following \cite{wang2019zero,lu2019see}, we adopt two alternated steps to train our model. In the static-image iteration, we utilize image saliency dataset MSRA10K \cite{cheng2014global} to fine-tune the DeepLabV3 based feature embedding module and the Center Prediction Branch. This allows the backbone to extract more discriminative foreground features and the center prediction module to locate the center of salient object more accurately. Meanwhile, in the dynamic-video iteration, we train the whole model with the training set in DAVIS16 \cite{perazzi2016benchmark}. During training the Center Guiding Appearance Diffusion module, we first separately train the Center Prediction Branch and Spatial-Prior Guided Appearance Matching in parallel in the first 20 epochs, where we feed the appearance matching module with the center point of ground truth. Then, in latter epochs, we feed the appearance matching module with the predicted point from the Center Prediction Branch to jointly train them.
For the above alternated training process, the size of input RGB frame is $473 \times 473 \times 3$. The entire network is trained using the SGD optimizer with an initial learning rate of $2.5 \times10^{-4}$. We set the batchsize as 16. All the experiments are conducted using 4 V100 GPUs on a server. The overall training time is about 9 hours, \ldz{and it takes about 0.1s with one image in a forward pass.}


\begin{table}[t!]
    \centering
    \caption{Ablation study of the proposed model on DAVIS-2016 dataset, measured by Mean $\mathcal{J}$ and Mean $\mathcal{F}$.}
    \begin{tabular}{l||cc|cc}
    \hlinew{1pt}
    Network Variant & Mean $\mathcal{J} \uparrow$ & $\bigtriangleup \mathcal{J}$ & Mean $\mathcal{F} \uparrow$ & $\bigtriangleup \mathcal{F}$ \\ \hline \hline
    Baseline & 77.6 & -5.5 & 76.8 & -7.6 \\ \hline
    + AD & 79.6 & -3.5 & 79.0 & -5.4 \\
    + CGAD & 82.5 & -0.6 & 82.9 & -1.5 \\ \hline
    + CGAD\&SA & 82.7 & -0.4 & 83.4 & -1.0 \\
    + CGAD\&CA & 82.8 & -0.3 & 83.6 & -0.8 \\
    + CGAD\&SCA & 83.0 & -0.1 & 84.2 & -0.2 \\
    + CGAD\&CSA & \textbf{83.1} & - & \textbf{84.4} & - \\ \hlinew{1pt}
    \end{tabular}
    \label{tab:ablation1}
    \vspace{-12pt}
\end{table}


\begin{table}[t!]
    \centering
    \caption{Ablation study of different strategies to choose the center point from the predicted heatmap.}
    \begin{tabular}{c||cc}
    \hlinew{1pt}
    Mechanism & Mean $\mathcal{J} \uparrow$ & Mean $\mathcal{F} \uparrow$ \\ \hline \hline
    maximum & 82.4 & 84.2 \\
    motion based \cite{xu2019mhp} & \textbf{83.1} & \textbf{84.4} \\ \hlinew{1pt}
    \end{tabular}
    \label{tab:select}
    \vspace{-12pt}
\end{table}

\begin{table*}[t!]
    \centering
    \caption{Quantitative results of UVOS methods on the DAVIS2016 validation set. All the results are borrowed from the public leaderboard maintained by the DAVIS challenge. The best scores are marked in \textbf{bold}.}
    \begin{tabular}{p{0.5cm}<{\centering}c||ccccccccccccc}
    \hlinew{1pt}
    ~ & Method & FSEG & UOVOS & LVO & ARP & PDB & LSMO & MoA & EpO & AGS & COS & AnDiff & MAT & Ours\\ \hline \hline
    $\mathcal{J}$\&$\mathcal{F}$ & Mean$\uparrow$ & 68.0 & 70.9 & 74.0 & 73.4 & 75.9 & 77.1 & 77.3 & 78.1 & 78.6 & 80.0 & 81.1 & 81.5 & \textbf{83.7} \\ \hline
    \multirow{3}*{$\mathcal{J}$} & Mean$\uparrow$ & 70.7 & 73.9 & 75.9 & 76.2 & 77.2 & 78.2 & 77.2 & 80.6 & 79.7 & 80.5 & 81.7 & 82.4 & \textbf{83.1} \\
    ~ & Recall$\uparrow$ & 83.5 & 88.5 & 89.1 & 91.1 & 90.1 & 89.1 & 87.8 & 95.2 & 91.1 & 93.1 & 90.9 & 94.5 & \textbf{95.7} \\
    ~ & Decay$\downarrow$ & 1.5 & 0.6 & \textbf{0.0} & 7.0 & 0.9 & 4.1 & 5.0 & 2.2 & 1.9 & 4.4 & 2.2 & 5.5 & \textbf{0.0}\\ \hline 
    \multirow{3}*{$\mathcal{F}$} & Mean$\uparrow$ & 65.3 & 68.0 & 72.1 & 70.6 & 74.5 & 75.9 & 77.4 & 75.5 & 77.4 & 79.5 & 80.5 & 80.7 & \textbf{84.4} \\
    ~ & Recall$\uparrow$ & 73.8 & 80.6 & 83.4 & 83.5 & 84.4 & 84.7 & 84.4 & 87.9 & 85.8 & 89.5 & 85.1 & 90.2 & \textbf{92.3} \\
    ~ & Decay$\downarrow$ & 1.8 & 0.7 & 1.3 & 7.9 & \textbf{-0.2} & 3.5 & 3.3 & 2.4 & 1.6 & 5.0 & 0.6 & 4.5 & 0.8 \\ \hline 
    $\mathcal{T}$ & Mean$\downarrow$ & 32.8 & 39.0 & 26.5 & 39.3 & 29.1 & 21.2 & 27.9 & 19.3 & 26.7 & \textbf{18.4} & 21.4 & 21.6 & 20.9 \\ \hlinew{1pt}
    \end{tabular}
    \label{tab:davis}
    \vspace{-12pt}
\end{table*}

\section{Experiments}
\subsection{Experimental Setup}
We conduct experiments on three well-known datasets: DAVIS2016 \cite{perazzi2016benchmark}, Youtube-Objects \cite{prest2012learning}, and FBMS \cite{ochs2013segmentation}.

\noindent \textbf{DAVIS2016} is a challenging video object segmentation dataset which consists of 50 videos in total (30 for training and 20 for validation) with pixel-wise annotations for every frame. Three evaluation criteria are used following the standard evaluation protocol \cite{perazzi2016benchmark}: region similarity $\mathcal{J}$, boundary accuracy $\mathcal{F}$, and time stability $\mathcal{T}$.

\noindent \textbf{Youtube-Objects} contains 126 video sequences which belong to 10 objects categories with more than 20,000 frames in total. Following its protocol, we use the region similarity $\mathcal{J}$ to measure the segmentation performance.

\noindent \textbf{FBMS} is comprised of 59 video sequences (29 training videos and 30 test videos). The ground-truth of FBMS is sparsely labeled. Following \cite{yang2019anchor,lu2019see,zhou2020motion}, we use region similarity $\mathcal{J}$ to evaluate our method on test set without training.


\subsection{Ablation Study}
We conduct the ablation study on DAVIS2016 dataset as shown in Table \ref{tab:ablation1} and \ref{tab:select}, where the baseline model is a the DeepLabV3 network.

\noindent \textbf{How does the gauss based spatial guidance help?}
We first investigate the effectiveness of the proposed Center Guiding Appearance Diffusion (CGAD) Module, which relies on the center point estimated by the Center Prediction Branch. We denote AD as normally appearance diffusion (AnDiff without multi-scale and pruning strategies), and CGAD as the center-prior guided one. Compare to the baseline model, as shown in Table \ref{tab:ablation1}, AD brings the improvement of 2.0 on $\mathcal{J}$ and 2.2 on $\mathcal{F}$ which indicates the effectiveness of the general appearance matching based method. Compare to it, our CGAD outperforms AD by 2.9 on $\mathcal{J}$ and 3.9 on $\mathcal{F}$ with a large margin. It demonstrates the effectiveness of the gauss based spatial prior, which helps to focus on the foreground and thus alleviates the challenges in common appearance matching based methods.

\noindent \textbf{How does the dynamic information fusion help?} 
The Dynamic Information Fusion Module consists of channel-wise and spatial-wise attention mechanisms. Here, we do the comparison on four variants: spatial attention only (SA), channel attention only (CA), first spatial attention then channel attention (SCA), and first channel attention then spatial attention (CSA). Detailed experiments on different variant attentions can be found in Table \ref{tab:ablation1}.
Compared to CGAD, SA can bring improvements of 0.2 on $\mathcal{J}$, 0.5 on $\mathcal{F}$, and CA can bring improvements of 0.3 on $\mathcal{J}$, 0.7 on $\mathcal{F}$. 
To investigate the performance on different combinations of the two attentions, we conduct the experiments on SCA and CSA. The results show that the channel-spatial attention (CSA) achieves the best performance of $\mathcal{J}=83.1$ and $\mathcal{F}=84.4$.

\noindent \textbf{How to select the center point from a heatmap?} 
Besides, we also do the ablation study on different strategies to choose the center point from the heatmap $\bm{H}$ in Eq. (\ref{eq:heatmap}). Specially, we compare two methods in Table \ref{tab:select}: 1) We directly choose the the center point in heatmap with the maximum score. 2) We exploit a motion mechanism \cite{xu2019mhp} to choose the center point with motion history information under the parameter settings $K=5,n=10$. We can find that the motion mechanism performs relatively better. 
It reveals that
the motion mechanism can help locate more accurate position of object based on the motion history,
whether objects move fast or slow in the video.

\begin{table}[t!]
\small
    \centering
    \caption{Quantitative performance of each category on Youtube-Objects with the Mean $\mathcal{J}$.}
    \begin{tabular}{c||p{0.5cm}p{0.5cm}p{0.5cm}p{0.5cm}p{0.5cm}p{0.5cm}p{0.5cm}p{0.5cm}}
    \hlinew{1pt}
    Method & FSEG & LVO & PDB & AGS & COS & MAT & Ours \\ \hline \hline
    Airplane & 81.7 & \textbf{86.2} & 78.0 & 87.7 & 81.1 & 72.9 & 85.8 \\
    Bird & 63.8 & 81.0 & 80.0 & 76.7 & 75.7 & 77.5 & \textbf{82.8} \\
    Boat & 72.3 & 68.5 & 58.9 & 72.2 & 71.3 & 66.9 & \textbf{81.9} \\
    Car & 74.9 & 69.3 & 76.5 & 78.6 & 77.6 & 79.0 & \textbf{81.4} \\
    Cat & 68.4 & 58.8 & 63.0 & 69.2 & 66.5 & \textbf{73.7} & 70.2 \\
    Cow & 68.0 & 68.5 & 64.1 & 64.6 & 69.8 & 67.4 & \textbf{71.0} \\
    Dog & 69.4 & 61.7 & 70.1 & 73.3 & \textbf{76.8} & 75.9 & 75.8 \\
    Horse & 60.4 & 53.9 & 67.6 & 64.4 & 67.4 & 63.2 & \textbf{75.4} \\
    Motorbike & 62.7 & 60.8 & 58.4 & 62.1 & 67.7 & 62.6 & \textbf{71.8} \\
    Train & 62.2 & \textbf{66.3} & 35.3 & 48.2 & 46.8 & 51.0 & 59.6 \\ \hline
    Mean$\mathcal{J}\uparrow$ & 68.4 & 67.5 & 65.5 & 69.7 & 70.5 & 69.0 & \textbf{75.6} \\ \hlinew{1pt}
    \end{tabular}
    \label{tab:youtube}
    \vspace{-12pt}
\end{table}

\begin{figure*}[t!]
\centering
\includegraphics[width=1.0\textwidth]{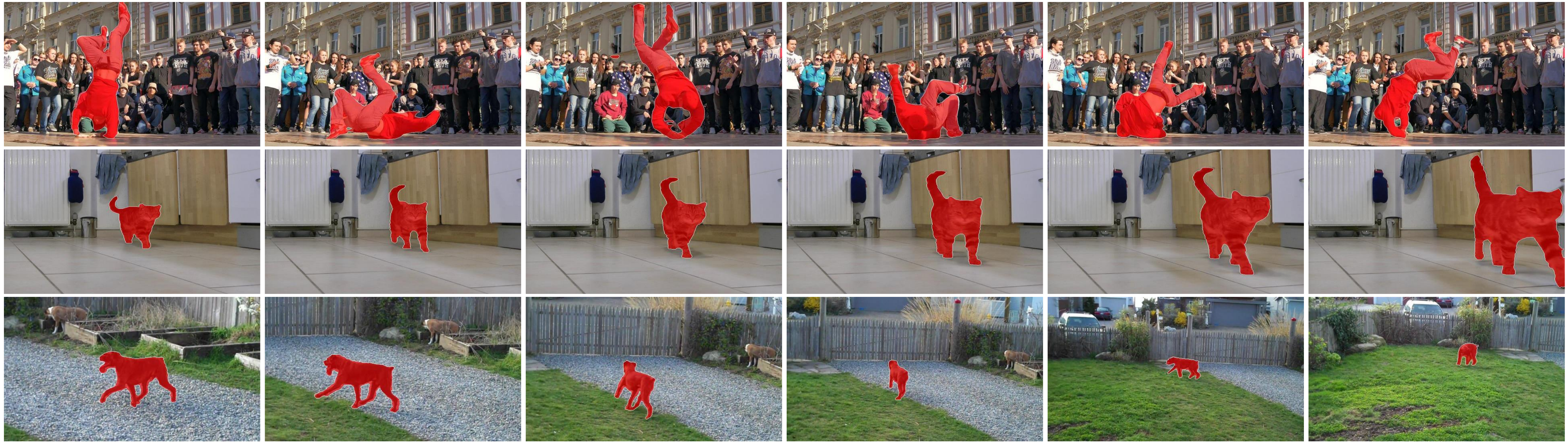}
\caption{Qualitative results on three datasets. From top to bottom: \textit{breakdance} from the DAVIS2016 dataset, \textit{cats01} from the
FBMS dataset, and \textit{dog0006} from the Youtube-Objects dataset.}
\label{fig:result1}
\vspace{-12pt}
\end{figure*}

\subsection{Comparison with State-of-the-arts}

\noindent \textbf{Evaluation on DAVIS2016.}
We compare our \textbf{F2Net} with the top performing UVOS methods in the public leaderboard on DAVIS2016 dataset. As shown in Table \ref{tab:davis}, our method outperforms all the reported methods across most metrics.  Compared with the second best method MAT \cite{zhou2020motion}, our model achieves gains of 2.2 in terms of $\mathcal{J}$\&$\mathcal{F}$ Mean. In detail, we obtains improvements of 0.7 and 3.7 on $\mathcal{J}$ Mean and $\mathcal{F}$ Mean, respectively.
Compared to appearance matching methods COS \cite{lu2019see} and AnDiff \cite{yang2019anchor}, we outperform them both on $\mathcal{J}$ Mean and $\mathcal{F}$ Mean by a large margin.
Compared to methods like MoA \cite{siam2019video} and EPO \cite{faisal2019exploiting} which utilize both appearance information and motion cues, our model outperforms them by only utilizing appearance information.
In our experiments, we find that the above methods may fail to distinguish the visually similar pixels from both foreground and background.
Due to the center based gauss map, our model consider spatial prior during the appearance matching procedure, which helps filter out the background noise and focus more on the foreground object boundary construction (improving $\mathcal{F}$). 
Besides, the Dynamic Information Fusion Module also can aggregate discriminative features across different level for the final segmentation.

\noindent \textbf{Evaluation on Youtube-Objects.}
Table \ref{tab:youtube} illustrates the results of all compared methods for different categories on Youtube-Objects dataset. Our approach brings improvement of 5.1 on $\mathcal{J}$ than the second best method COS \cite{lu2019see} by a large margin. It is also worth to note that we outperform all compared methods on almost all categories. The main reason lies in two folds: First, for optical guided methods MAT, FSEG \cite{jain2017fusionseg} and LVO \cite{tokmakov2017learning}, sequences in the Airplane
and Boat categories contain objects that have quick appearance variation or move slowly. Both factors result in inaccurate estimation of optical flow.
Compared to them, our matching based framework can handle
these scenarios well. Second, compared to other matching based methods COS, our estimated center point provides the spatial prior for better segmentation focusing more on the actual foreground.

\begin{table}[t!]
\small
    \centering
    \caption{Quantitative results on FBMS test set over Mean $\mathcal{J}$.}
    \begin{tabular}{c||cccccc}
    \hlinew{1pt}
    Method & NLC & FST & SFL & APR & MSTP & FSEG \\ \hline
    Mean$\mathcal{J}\uparrow$ & 44.5 & 55.5 & 56.0 & 59.8 & 60.8 & 68.4 \\ \hline \hline
    Method & IET & OBN & PDB & COS & MAT & Ours \\ \hline
    Mean$\mathcal{J}\uparrow$ & 71.9 & 73.9 & 74.0 & 75.6 & 76.1 & \textbf{77.5}\\ \hlinew{1pt}
    \end{tabular}
    \label{tab:fbms}
    \vspace{-12pt}
\end{table}

\noindent \textbf{Evaluation on FBMS.}
As shown in Table \ref{tab:fbms}, we also conduct experiments on FBMS for completeness. Compared to others, our method performs the best result with 77.5 over the Mean $\mathcal{J}$, outperforming the second best one by 1.4. Considering lots of foreground objects in FBMS share similar appearance with the background, our Center Guiding Appearance Diffusion Module exploits center information to focus on the foreground objects and filter out the visually similar background ones for better segmentation.

\subsection{Qualitative Results}
\begin{figure}[t!]
\centering
\includegraphics[width=0.48\textwidth]{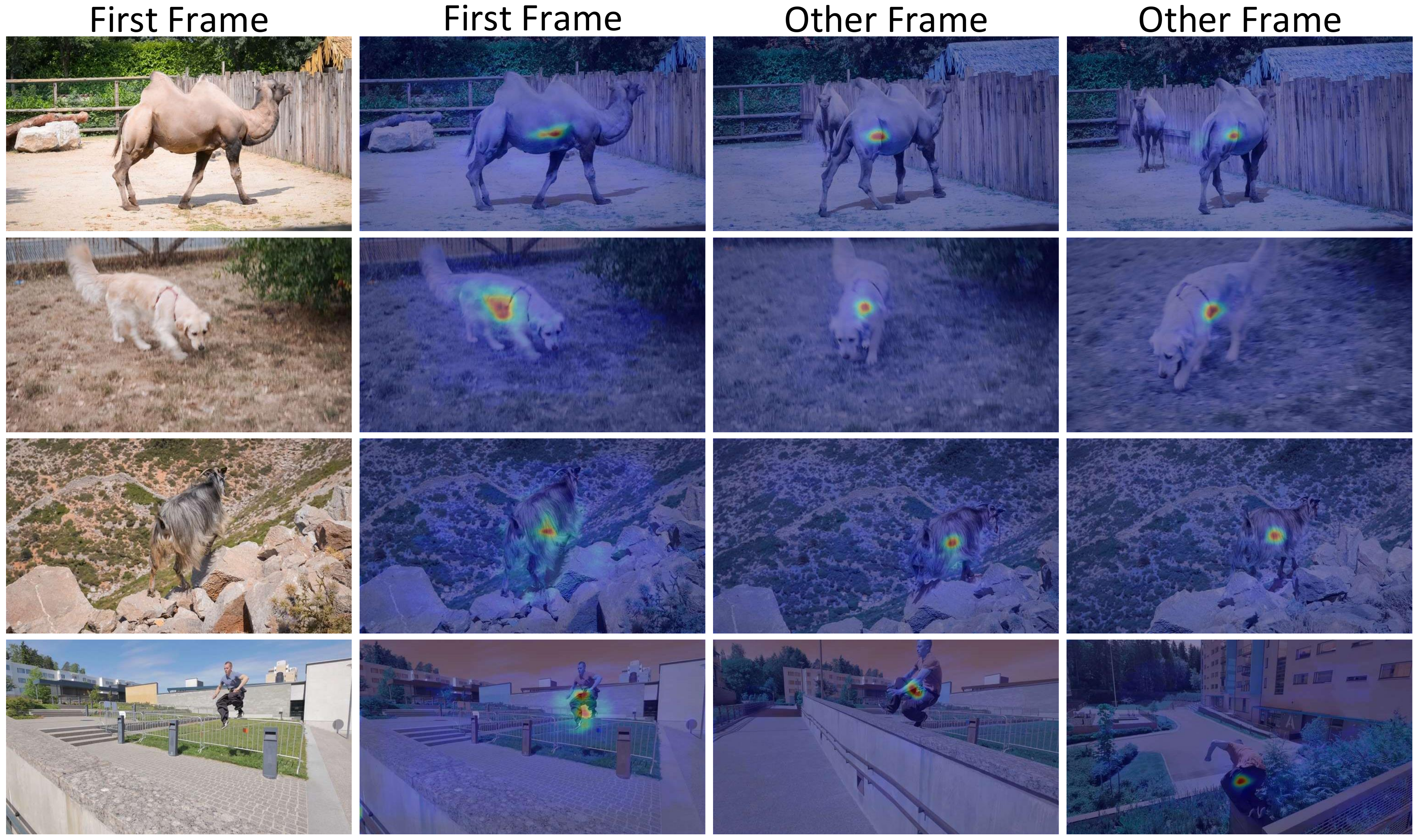}
\caption{Visualization of center point heatmaps on each frame of the DAVIS2016 dataset. From top to down: \textit{camel}, \textit{dog}, \textit{goat}, and \textit{parkour}. Although the first frame has no history motion knowledge of the foreground object, our Center Prediction Branch can still estimate accurate point result among noisy neighbors. The other frames receive additional point information from previous frame and thus predict more accurate center point.}
\label{fig:result2}
\vspace{-12pt}
\end{figure}

\noindent \textbf{Does center prediction branch estimate the center point well?}
To investigate the performance of Center Prediction Branch, we give some visualization results on the generated center point heatmap, especially for the first frame of each video sequence. As shown in Figure \ref{fig:result2}, there are four challenging sequences in which the surroundings has a similar appearance to the foreground object. Without any previous motion history, our Center Prediction Branch still can estimate relatively accurate object center point of the first frame mainly based on the semantic features. It demonstrates that our Center Prediction Branch can effectively capture the contextual information across the frame to locate the target point. In the latter frames, besides the current semantic features, the center point based gauss map from previous frame is additionally fed to refine the point position, which provides more precisely center point estimation.

\noindent \textbf{Segmentation visualization.} Figure \ref{fig:result1} shows qualitative results sampled from the three datasets. The \textit{breakdance} sequence from DAVIS2016 contains many challenging factors, such as fast motion, deformation and multiple instances of the same category. We can find that our method is robust to such complex scenarios and can accurately segment out primary objects from the cluttered background. The effectiveness is further proved in the \textit{cat01} sequence of FBMS dataset. In addition, our method performs well in the \textit{dog0006} sequence of Youtube-Objects, in which the target suffers from large scale variations.

\section{Conclusion}
In this paper, we propose a novel \textit{Focus on Foreground Network} (\textbf{F2Net}) for unsupervised video object segmentation. Compared to recent appearance matching based methods, we additionally estimate the center point of the foreground object and encode it into a gauss map as spatial guidance for the appearance matching procedure. This Center Guiding Appearance Diffusion Module is flexible and can be easily adapted to other segmentation frameworks. We also develop a Dynamic Information Fusion Module to aggregate multi-level features for generating more discriminative features for final segmentation. Extensive results on three public datasets demonstrate the effectiveness of our proposed method.

\clearpage
\noindent \ldz{\textbf{Acknowledgements.} This work was performed while Daizong Liu worked as an intern at ByteDance AI Lab. It was supported in part by the National Natural Science Foundation of China under Grant No.61972448.}

\bibliography{reference.bib}

\begin{thebibliography}{}

\bibitem[\protect\citeauthoryear{Chen \bgroup et al\mbox.\egroup
  }{2017}]{chen2017rethinking}
Chen, L.-C.; Papandreou, G.; Schroff, F.; and Adam, H.
\newblock 2017.
\newblock Rethinking atrous convolution for semantic image segmentation.
\newblock {\em arXiv preprint arXiv:1706.05587}.

\bibitem[\protect\citeauthoryear{Chen \bgroup et al\mbox.\egroup
  }{2018}]{chen2018blazingly}
Chen, Y.; Pont-Tuset, J.; Montes, A.; and Van~Gool, L.
\newblock 2018.
\newblock Blazingly fast video object segmentation with pixel-wise metric
  learning.
\newblock In {\em CVPR},  1189--1198.

\bibitem[\protect\citeauthoryear{Cheng \bgroup et al\mbox.\egroup
  }{2014}]{cheng2014global}
Cheng, M.-M.; Mitra, N.~J.; Huang, X.; Torr, P.~H.; and Hu, S.-M.
\newblock 2014.
\newblock Global contrast based salient region detection.
\newblock {\em TPAMI} 37(3):569--582.

\bibitem[\protect\citeauthoryear{Cheng \bgroup et al\mbox.\egroup
  }{2017}]{cheng2017segflow}
Cheng, J.; Tsai, Y.-H.; Wang, S.; and Yang, M.-H.
\newblock 2017.
\newblock Segflow: Joint learning for video object segmentation and optical
  flow.
\newblock In {\em ICCV},  686--695.

\bibitem[\protect\citeauthoryear{Chu \bgroup et al\mbox.\egroup
  }{2017}]{chu2017multi}
Chu, X.; Yang, W.; Ouyang, W.; Ma, C.; Yuille, A.~L.; and Wang, X.
\newblock 2017.
\newblock Multi-context attention for human pose estimation.
\newblock In {\em Proceedings of the IEEE Conference on Computer Vision and
  Pattern Recognition (CVPR)},  1831--1840.

\bibitem[\protect\citeauthoryear{Faisal \bgroup et al\mbox.\egroup
  }{2019}]{faisal2019exploiting}
Faisal, M.; Akhter, I.; Ali, M.; and Hartley, R.
\newblock 2019.
\newblock Exploiting geometric constraints on dense trajectories for motion
  saliency.
\newblock In {\em WACV}.

\bibitem[\protect\citeauthoryear{Faktor and Irani}{2014}]{faktor2014video}
Faktor, A., and Irani, M.
\newblock 2014.
\newblock Video segmentation by non-local consensus voting.
\newblock In {\em BMVC}, volume~2, ~8.

\bibitem[\protect\citeauthoryear{Fathi \bgroup et al\mbox.\egroup
  }{2017}]{fathi2017semantic}
Fathi, A.; Wojna, Z.; Rathod, V.; Wang, P.; Song, H.~O.; Guadarrama, S.; and
  Murphy, K.~P.
\newblock 2017.
\newblock Semantic instance segmentation via deep metric learning.
\newblock {\em arXiv preprint arXiv:1703.10277}.

\bibitem[\protect\citeauthoryear{He \bgroup et al\mbox.\egroup
  }{2016}]{he2016deep}
He, K.; Zhang, X.; Ren, S.; and Sun, J.
\newblock 2016.
\newblock Deep residual learning for image recognition.
\newblock In {\em CVPR},  770--778.

\bibitem[\protect\citeauthoryear{Hu, Huang, and
  Schwing}{2018}]{hu2018unsupervised}
Hu, Y.-T.; Huang, J.-B.; and Schwing, A.~G.
\newblock 2018.
\newblock Unsupervised video object segmentation using motion saliency-guided
  spatio-temporal propagation.
\newblock In {\em ECCV},  786--802.

\bibitem[\protect\citeauthoryear{Hu, Shen, and Sun}{2018}]{hu2018squeeze}
Hu, J.; Shen, L.; and Sun, G.
\newblock 2018.
\newblock Squeeze-and-excitation networks.
\newblock In {\em CVPR},  7132--7141.

\bibitem[\protect\citeauthoryear{Jain, Xiong, and
  Grauman}{2017}]{jain2017fusionseg}
Jain, S.~D.; Xiong, B.; and Grauman, K.
\newblock 2017.
\newblock Fusionseg: Learning to combine motion and appearance for fully
  automatic segmentation of generic objects in videos.
\newblock In {\em CVPR},  2117--2126.

\bibitem[\protect\citeauthoryear{Law and Deng}{2018}]{law2018cornernet}
Law, H., and Deng, J.
\newblock 2018.
\newblock Cornernet: Detecting objects as paired keypoints.
\newblock In {\em ECCV},  734--750.

\bibitem[\protect\citeauthoryear{Li \bgroup et al\mbox.\egroup
  }{2018a}]{li2018flow}
Li, G.; Xie, Y.; Wei, T.; Wang, K.; and Lin, L.
\newblock 2018a.
\newblock Flow guided recurrent neural encoder for video salient object
  detection.
\newblock In {\em CVPR},  3243--3252.

\bibitem[\protect\citeauthoryear{Li \bgroup et al\mbox.\egroup
  }{2018b}]{li2018instance}
Li, S.; Seybold, B.; Vorobyov, A.; Fathi, A.; Huang, Q.; and Jay~Kuo, C.-C.
\newblock 2018b.
\newblock Instance embedding transfer to unsupervised video object
  segmentation.
\newblock In {\em CVPR},  6526--6535.

\bibitem[\protect\citeauthoryear{Li \bgroup et al\mbox.\egroup
  }{2018c}]{li2018unsupervised}
Li, S.; Seybold, B.; Vorobyov, A.; Lei, X.; and Jay~Kuo, C.-C.
\newblock 2018c.
\newblock Unsupervised video object segmentation with motion-based bilateral
  networks.
\newblock In {\em ECCV},  207--223.

\bibitem[\protect\citeauthoryear{Li \bgroup et al\mbox.\egroup
  }{2019}]{li2019selective}
Li, X.; Wang, W.; Hu, X.; and Yang, J.
\newblock 2019.
\newblock Selective kernel networks.
\newblock In {\em Proceedings of the IEEE Conference on Computer Vision and
  Pattern Recognition (CVPR)},  510--519.

\bibitem[\protect\citeauthoryear{Lin \bgroup et al\mbox.\egroup
  }{2017}]{lin2017focal}
Lin, T.-Y.; Goyal, P.; Girshick, R.; He, K.; and Doll{\'a}r, P.
\newblock 2017.
\newblock Focal loss for dense object detection.
\newblock In {\em ICCV},  2980--2988.

\bibitem[\protect\citeauthoryear{Lu \bgroup et al\mbox.\egroup
  }{2016}]{lu2016hierarchical}
Lu, J.; Yang, J.; Batra, D.; and Parikh, D.
\newblock 2016.
\newblock Hierarchical question-image co-attention for visual question
  answering.
\newblock In {\em Advances in Neural Information Processing Systems (NIPS)},
  289--297.

\bibitem[\protect\citeauthoryear{Lu \bgroup et al\mbox.\egroup
  }{2019}]{lu2019see}
Lu, X.; Wang, W.; Ma, C.; Shen, J.; Shao, L.; and Porikli, F.
\newblock 2019.
\newblock See more, know more: Unsupervised video object segmentation with
  co-attention siamese networks.
\newblock In {\em CVPR},  3623--3632.

\bibitem[\protect\citeauthoryear{Ochs and Brox}{2011}]{ochs2011object}
Ochs, P., and Brox, T.
\newblock 2011.
\newblock Object segmentation in video: a hierarchical variational approach for
  turning point trajectories into dense regions.
\newblock In {\em ICCV},  1583--1590.

\bibitem[\protect\citeauthoryear{Ochs, Malik, and
  Brox}{2013}]{ochs2013segmentation}
Ochs, P.; Malik, J.; and Brox, T.
\newblock 2013.
\newblock Segmentation of moving objects by long term video analysis.
\newblock {\em TPAMI} 36(6):1187--1200.

\bibitem[\protect\citeauthoryear{Oh \bgroup et al\mbox.\egroup
  }{2019}]{oh2019video}
Oh, S.~W.; Lee, J.-Y.; Xu, N.; and Kim, S.~J.
\newblock 2019.
\newblock Video object segmentation using space-time memory networks.
\newblock In {\em Proceedings of the IEEE International Conference on Computer
  Vision (ICCV)},  9226--9235.

\bibitem[\protect\citeauthoryear{Papazoglou and
  Ferrari}{2013}]{papazoglou2013fast}
Papazoglou, A., and Ferrari, V.
\newblock 2013.
\newblock Fast object segmentation in unconstrained video.
\newblock In {\em ICCV},  1777--1784.

\bibitem[\protect\citeauthoryear{Perazzi \bgroup et al\mbox.\egroup
  }{2016}]{perazzi2016benchmark}
Perazzi, F.; Pont-Tuset, J.; McWilliams, B.; Van~Gool, L.; Gross, M.; and
  Sorkine-Hornung, A.
\newblock 2016.
\newblock A benchmark dataset and evaluation methodology for video object
  segmentation.
\newblock In {\em CVPR},  724--732.

\bibitem[\protect\citeauthoryear{Prest \bgroup et al\mbox.\egroup
  }{2012}]{prest2012learning}
Prest, A.; Leistner, C.; Civera, J.; Schmid, C.; and Ferrari, V.
\newblock 2012.
\newblock Learning object class detectors from weakly annotated video.
\newblock In {\em CVPR},  3282--3289.

\bibitem[\protect\citeauthoryear{Siam \bgroup et al\mbox.\egroup
  }{2019}]{siam2019video}
Siam, M.; Jiang, C.; Lu, S.; Petrich, L.; Gamal, M.; Elhoseiny, M.; and
  Jagersand, M.
\newblock 2019.
\newblock Video object segmentation using teacher-student adaptation in a human
  robot interaction (hri) setting.
\newblock In {\em ICRA},  50--56.

\bibitem[\protect\citeauthoryear{Su \bgroup et al\mbox.\egroup
  }{2019}]{su2019multi}
Su, K.; Yu, D.; Xu, Z.; Geng, X.; and Wang, C.
\newblock 2019.
\newblock Multi-person pose estimation with enhanced channel-wise and spatial
  information.
\newblock In {\em Proceedings of the IEEE Conference on Computer Vision and
  Pattern Recognition (CVPR)},  5674--5682.

\bibitem[\protect\citeauthoryear{Tokmakov, Alahari, and
  Schmid}{2017}]{tokmakov2017learning}
Tokmakov, P.; Alahari, K.; and Schmid, C.
\newblock 2017.
\newblock Learning motion patterns in videos.
\newblock In {\em CVPR},  3386--3394.

\bibitem[\protect\citeauthoryear{Tompson \bgroup et al\mbox.\egroup
  }{2014}]{tompson2014joint}
Tompson, J.~J.; Jain, A.; LeCun, Y.; and Bregler, C.
\newblock 2014.
\newblock Joint training of a convolutional network and a graphical model for
  human pose estimation.
\newblock In {\em NIPS},  1799--1807.

\bibitem[\protect\citeauthoryear{Tsai, Yang, and Black}{2016}]{tsai2016video}
Tsai, Y.-H.; Yang, M.-H.; and Black, M.~J.
\newblock 2016.
\newblock Video segmentation via object flow.
\newblock In {\em CVPR},  3899--3908.

\bibitem[\protect\citeauthoryear{Wang \bgroup et al\mbox.\egroup
  }{2018}]{wang2018non}
Wang, X.; Girshick, R.; Gupta, A.; and He, K.
\newblock 2018.
\newblock Non-local neural networks.
\newblock In {\em CVPR},  7794--7803.

\bibitem[\protect\citeauthoryear{Wang \bgroup et al\mbox.\egroup
  }{2019}]{wang2019zero}
Wang, W.; Lu, X.; Shen, J.; Crandall, D.~J.; and Shao, L.
\newblock 2019.
\newblock Zero-shot video object segmentation via attentive graph neural
  networks.
\newblock In {\em ICCV},  9236--9245.

\bibitem[\protect\citeauthoryear{Wang \bgroup et al\mbox.\egroup
  }{2020}]{wang2020centermask}
Wang, Y.; Xu, Z.; Shen, H.; Cheng, B.; and Yang, L.
\newblock 2020.
\newblock Centermask: single shot instance segmentation with point
  representation.
\newblock In {\em CVPR},  9313--9321.

\bibitem[\protect\citeauthoryear{Wang, Shen, and
  Porikli}{2015}]{wang2015saliency}
Wang, W.; Shen, J.; and Porikli, F.
\newblock 2015.
\newblock Saliency-aware geodesic video object segmentation.
\newblock In {\em CVPR},  3395--3402.

\bibitem[\protect\citeauthoryear{Xu \bgroup et al\mbox.\egroup
  }{2019}]{xu2019mhp}
Xu, S.; Liu, D.; Bao, L.; Liu, W.; and Zhou, P.
\newblock 2019.
\newblock Mhp-vos: Multiple hypotheses propagation for video object
  segmentation.
\newblock In {\em CVPR},  314--323.

\bibitem[\protect\citeauthoryear{Yang \bgroup et al\mbox.\egroup
  }{2018}]{yang2018efficient}
Yang, L.; Wang, Y.; Xiong, X.; Yang, J.; and Katsaggelos, A.~K.
\newblock 2018.
\newblock Efficient video object segmentation via network modulation.
\newblock In {\em CVPR},  6499--6507.

\bibitem[\protect\citeauthoryear{Yang \bgroup et al\mbox.\egroup
  }{2019}]{yang2019anchor}
Yang, Z.; Wang, Q.; Bertinetto, L.; Hu, W.; Bai, S.; and Torr, P.~H.
\newblock 2019.
\newblock Anchor diffusion for unsupervised video object segmentation.
\newblock In {\em ICCV},  931--940.

\bibitem[\protect\citeauthoryear{Zhang, Javed, and Shah}{2013}]{zhang2013video}
Zhang, D.; Javed, O.; and Shah, M.
\newblock 2013.
\newblock Video object segmentation through spatially accurate and temporally
  dense extraction of primary object regions.
\newblock In {\em CVPR},  628--635.

\bibitem[\protect\citeauthoryear{Zhou \bgroup et al\mbox.\egroup
  }{2020}]{zhou2020motion}
Zhou, T.; Wang, S.; Zhou, Y.; Yao, Y.; Li, J.; and Shao, L.
\newblock 2020.
\newblock Motion-attentive transition for zero-shot video object segmentation.
\newblock In {\em AAAI}, volume~2, ~3.

\bibitem[\protect\citeauthoryear{Zhou, Koltun, and
  Kr{\"a}henb{\"u}hl}{2020}]{zhou2020tracking}
Zhou, X.; Koltun, V.; and Kr{\"a}henb{\"u}hl, P.
\newblock 2020.
\newblock Tracking objects as points.
\newblock {\em arXiv:2004.01177}.

\bibitem[\protect\citeauthoryear{Zhou, Wang, and
  Kr{\"a}henb{\"u}hl}{2019}]{zhou2019objects}
Zhou, X.; Wang, D.; and Kr{\"a}henb{\"u}hl, P.
\newblock 2019.
\newblock Objects as points.
\newblock {\em arXiv preprint arXiv:1904.07850}.

\end{thebibliography}
\bibliographystyle{aaai}

\end{document}